\pdfoutput=1

\documentclass[11pt]{article}
\usepackage{acl}

\usepackage{times}
\usepackage{latexsym}
\usepackage[T1]{fontenc}
\usepackage[utf8]{inputenc}
\usepackage{microtype}
\usepackage{inconsolata}

\usepackage{amsfonts}
\usepackage{amsmath}
\usepackage{amssymb}
\usepackage{graphicx}
\usepackage{booktabs}
\usepackage{multirow}
\usepackage{xcolor, soul}

\usepackage{enumitem}

\AtBeginDocument{%
  \addtolength\abovedisplayskip{-0.3\baselineskip}%
  \addtolength\belowdisplayskip{-0.3\baselineskip}%
}

\newtheorem{definition}{Definition}

\newtheorem{theorem}{Theorem}

\newcommand\mask{\texttt{<mask>}}
\newcommand\be{\mathbf{e}}
\newcommand\bs{\mathbf{s}}
\newcommand\bx{\mathbf{x}}
\newcommand\ppl{\text{PPL}}
\newcommand\Dtrain{D}

\newcommand\spa{$\rho$\xspace}
\newcommand\spat{$\rho^*$\xspace}
\newcommand\ap{$\kappa$\xspace}

\newcommand\defense{SP-Defense\xspace}
\newcommand\deshort{SP-D\xspace}
\newcommand\attack{SP-Attack\xspace}

\newcommand\alg{EUBA\xspace}
\usepackage{color}

\usepackage[ruled,vlined,linesnumbered]{algorithm2e}

\SetCommentSty{mycommfont}

\title{Single Word Change is All You Need:\\
Using LLMs to Create Synthetic Training Examples for Text Classifiers}

\newcommand*{\affaddr}[1]{#1} 
\newcommand*{\affmark}[1][*]{\textsuperscript{#1}}
\newcommand*{\email}[1]{\texttt{#1}}

\author{
Lei Xu\affmark[1], Sarah Alnegheimish\affmark[1], Laure Berti-Equille\affmark[2],\\
\textbf{Alfredo Cuesta-Infante\affmark[3], Kalyan Veeramachaneni\affmark[1]} \\
\affaddr{\affmark[1] MIT LIDS}  \hspace{1ex}
\affaddr{\affmark[2] IRD  \hspace{1ex}
\affaddr{\affmark[3] Universidad Rey Juan Carlos}}\\
\email{\{leix,smish,kalyanv\}@mit.edu, laure.berti@ird.fr, alfredo.cuesta@urjc.es}
}

\begin{document}
\maketitle

\begin{abstract}
In text classification, creating an adversarial example means subtly perturbing a few words in a sentence without changing its meaning, causing it to be misclassified by a classifier. 
A concerning observation is that a significant portion of adversarial examples generated by existing methods change only one word. This single-word perturbation vulnerability represents a significant weakness in classifiers, which malicious users can exploit to efficiently create a multitude of adversarial examples.
This paper studies this problem and makes the following key contributions:
(1)~We introduce a novel metric \spa to quantitatively assess a classifier's \textit{robustness against single-word perturbation}.
(2)~We present the \textit{\attack}, designed to exploit the single-word perturbation vulnerability, achieving a higher attack success rate, better preserving sentence meaning, while reducing computation costs compared to state-of-the-art adversarial methods.
(3)~We propose \textit{\defense}, which aims to improve \spa by applying data augmentation in learning. 
Experimental results on 4 datasets and {2 masked language models} show that \defense improves \spa by 14.6\% and 13.9\%  and decreases the attack success rate of \attack by 30.4\% and 21.2\% on two classifiers respectively, and decreases the attack success rate of existing attack methods that involve multiple-word perturbations. 
\end{abstract}

\section{Introduction}

With the proliferation of generative AI-based applications, text classifiers have become mission critical. These classifiers are designed to be guardrails around the outputs of applications built on large language models. A text classifier can be used to detect whether a chatbot is accidentally giving financial advice, or leaking sensitive information. They have also frequently been used for security-critical applications, such as misinformation detection~\cite{wu2019misinformation, torabi2019big, zhou2019fake}. 

At the same time, a steady stream of work has shown that these classifiers are vulnerable to simple perturbations -- subtly modified sentences that, while practically indistinguishable from the original sentences, nonetheless change the output label of a classifier, causing it to fail. 

Deploying these classifiers alongside generative AI applications has brought up the question of whether the failure modes of these classifiers can be identified before the classifiers are deployed -- and, if they can be identified, whether they can also be updated to overcome these failure modes. 

One promising solution is to synthetically construct these "subtly modified" sentences in order to test the classifiers, and also to create sentences we can add to the training data to make the classifiers more robust in the first place. Not surprisingly, synthetic data generation is also possible with large language models. In this paper, we propose and demonstrate the use of large language models to construct synthetic sentences that are practically indistinguishable from the real sentences. 

In classical literature, sentences that can trick these classifiers are known as "adversarial examples." The process of trying to find these examples for a given classifier (or in general) is called an "adversarial attack," and training the classifier to be robust to these attacks is called "defense." We will use these terms here, as they enable us to compare our method with other well-known methods using similar notations and language.

A common metric for assessing the vulnerability of a classifier to such attacks is the Attack Success Rate (ASR), defined as the percentage of labeled sentences for which an adversarial sentence could be successfully designed.
Recent proposed methods can achieve an ASR of over 80\%~\cite{Jin2019textfooler, li2021CLARE}, making adversarial vulnerability a severe security issue.

These attacks work through black box methods, accessing only the output of a classifier. 
They tweak sentences in an iterative fashion, changing one or multiple words and then querying the classifier for its output until an adversarial example is achieved. When we examined the adversarial sentences generated by these methods, we were surprised to find that in a significant portion of them, just one word had been changed. In Table~\ref{tab:intro-single-ratio} we show results of three existing methods, 
their attack success rates, and {the percentage of adversarial examples they generated that had only single-world changes (SP)}. For example, 66\% of adversarial examples generated by CLARE attacks~\cite{li2021CLARE} on a movie review (MR) dataset~\cite{pang2005seeing} changed only one word.
Furthermore, many different adversarial sentences used the same word, albeit placed in different positions. For example, inserting the word ``online'' into a sentence repeatedly caused business news articles to be misclassified as technology news articles.

\begin{table}[t]
\centering
\begingroup
\caption{The ASR and percentage of adversarial examples with single-word perturbation (denoted as SP\%). We attack a vanilla BERT classifier on AG and MR, the  datasets used in our experiments, using TextFooler~\cite{Jin2019textfooler}, BAE~\cite{garg2020bae}, CLARE~\cite{li2021CLARE}, and our \attack. }\label{tab:intro-single-ratio}
\setlength{\tabcolsep}{5pt}
\resizebox{\columnwidth}{!}{%
\begin{tabular}{rrrrrrrrr}
\toprule
 & \multicolumn{2}{c}{\textbf{TextFooler}} &  \multicolumn{2}{c}{\textbf{BAE}} & \multicolumn{2}{c}{\textbf{CLARE}} & \multicolumn{2}{c}{\textbf{\attack}}\\
 \cmidrule(lr){2-3}\cmidrule(lr){4-5}\cmidrule(lr){6-7} \cmidrule(lr){8-9}
 & \textbf{ASR} & \textbf{SP\%} & \textbf{ASR} & \textbf{SP\%} & \textbf{ASR} & \textbf{SP\%} & \textbf{ASR} & \textbf{SP\%} \\
\midrule
\textbf{AG} & 65.2 & 17.0 & 19.3 & 35.8 & 84.4 & 38.6 & 82.7 & 100\\
\textbf{MR} & 72.4 & 49.2 & 41.3 & 66.6 & 90.0 & 66.2 & 93.5 & 100 \\\bottomrule
\end{tabular}}
\endgroup
\end{table}

Directly modeling this vulnerability is crucial, as both of these properties are beneficial to attackers. First, the fewer words that are changed, the more semantically similar the adversarial example will be to the original, making it look more "innocent" and running the risk that it will not be detected by other methods.
Second, knowing that a particular word can reliably change the classification will enable an attacker to reduce the number of queries to the classifier in a black box attack, which is usually $l$ queries, where $l$ is the sentence length.

Despite the attractive properties of direct methods for attackers, to the best of our knowledge, no existing work has been done to design a single-word perturbation- based attack -- nor is there a specific metric to quantify the robustness of a classifier against single-word perturbations. We approach this problem in a comprehensive way.


First, we define a measure -- the \textit{single-word flip capability}, denoted as \ap -- for each word in the classifier's vocabulary. This measure, presented in Section~\ref{sec:def}, is the percentage of sentences whose classification can be successfully changed by replacing that word at one of the positions. A subset of such sentences that are fluent enough and similar enough to the original sentence would comprise legitimate attacks.  
\ap allows us to measure a text classifier's \textit{robustness against single-word perturbations}, denoted as \spa. \spa is the percentage of sentence-word pairs in the Cartesian product of the dataset and the vocabulary where the classifier will not flip regardless of the word's position, meaning a successful attack is not possible.

The vocabulary size of the classifiers and datasets we are working with are too large for us to compute these metrics through brute force (30k words in the classifier's vocabulary, and datasets with 10k sentences of 30 words on average).
Therefore, in Section~\ref{sec:alg}, we propose an \textit{efficient upper bound algorithm} (EUBA) for \spa, which works by finding as many successful attacks as possible within a given time budget. We use first-order Taylor approximation to estimate the potential for each single-word substitution to lead to a successful attack. By prioritizing substitutions with high potential, we can find most of the successful attacks through verifying a subset of word substitutions rather than all of them, reducing the number of queries to the classifier.

With these two measures, we next design an efficient algorithm, \attack, to attack a classifier using \textit{single-word perturbations} while still maintaining fluency and semantic similarity (Section~\ref{sec:attack}). \attack pre-computes \ap for all words using EUBA, then performs an attack by changing only one word in a sentence, switching it to another high-capacity word (i.e., high \ap) to trigger a misclassification.
We show that \attack is more efficient than existing attack methods due to pre-computing \ap and flipping only one word. 


To overcome this type of attack, in Section~\ref{sec:defense}, we design \defense, which leverages the first-order approximation to find single-word perturbations and augment the training set. We retrain the classifier to become more robust -- not only against single-word perturbation attacks, but also against attacks involving multiple word changes.

We also carry out extensive experiments {on masked language models, specifically BERT classifiers}, in Section~\ref{sec:exp} to show that: 1) \ap and \spa are necessary robustness metrics; 2) EUBA is an efficient way to avoid brute force when computing the robustness metric \spa; 3) \attack can match or improve ASR compared to existing attacks that change multiple words, and better preserve the semantic meanings; 4) \defense is effective in defending against baseline attacks as well as against \attack, thus improving classifier robustness.

\begin{table}[!tb]
\centering
\resizebox{\columnwidth}{!}{%
\begin{tabular}{lp{6.5cm}}
\toprule
\textbf{Notation}   & \textbf{Description} \\ \midrule
$f(\cdot)$          & A text classifier \\
$V$                 & The vocabulary used by $f$ \\
$D$                 & A text classification training set \\
$D^+$               & A subset of $D$ where $f$ can correctly classify \\
$D^*$               & The Cartesian product of $D^+$ and $V$ \\
$S(\bx, w)$         & A set of sentences constructed by replacing one word in $\bx$ with $w$\\
$\kappa_f(w)$       & The single-word adversarial capability of $w$\\
$\rho(f)$           & The robustness against single-word perturbation measured on the training set \\
$\rho^*(f)$         & The robustness against single-word perturbation measured on the test set\\
ASR                 & Attack success rate \\\bottomrule
\end{tabular}}
\caption{Our Notations.}\label{tab:notation}
\end{table}

\begin{figure*}[tb]
    \centering
    \vspace{-1em}
    \includegraphics[width=\textwidth]{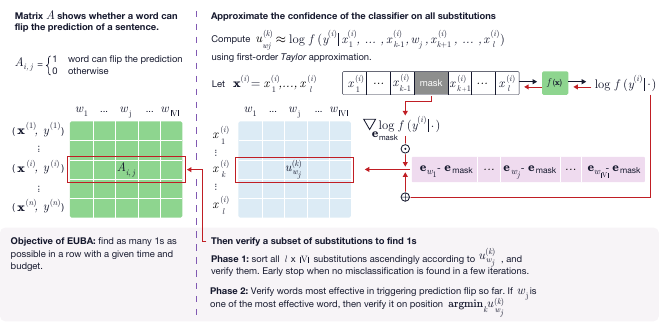}
    \vspace{-2em}
    \caption{The EUBA algorithm can efficiently find prediction flip by using first-order Taylor approximations. }\label{fig:euba}
    \vspace{-1em}
\end{figure*}

\section{Quantifying Classifier Robustness against Single-word Perturbations}\label{sec:def}
In this section, we formulate the single-word perturbation adversarial attack problem, and define two novel and useful metrics: \ap, which represents how capable a particular word is at flipping a prediction, and \spa, which represents the robustness of a text classifier against single-word perturbation. 
{We also refer the reader to Table} \ref{tab:notation}  {for other notations used throughout the paper.}

\subsection{Single-word Perturbation Attack Setup}
We consider a restricted adversarial attack scenario where the attacker can substitute only one word. The attack is considered successful if the classifier's prediction differs from the label. 

Let $D=(\bx_i, y_i)_{i=1\ldots |D|}$ be a text classification dataset, where $\bx_i$ is a sentence composed of words, $y_i$ is the label of the sentence and $|D|$ denotes the cardinality of $D$. 
A classifier $f(\cdot)$ is trained on $D$ to predict $y_i$ with input $\bx_i$. Additionally, $V$ is the vocabulary associated with $f(\cdot)$. 
The goal of the single-word perturbation attack is to construct an $\bx'$ for a sentence-label pair $(\bx, y)$ by replacing one word so that
\begin{itemize}
    \item the prediction is flipped, i.e., $f(\bx') \neq y$;
    \item $\bx'$ is a fluent sentence;
    \item $\bx'$ is semantically similar to $\bx$
\end{itemize}
Fluency of the sentence is measured by perplexity, while the similarity is measured by the cosine similarity of a Universal Sentence Encoder (USE)~\cite{cer2018universal} and human annotators.


\subsection{Two Useful Metrics For Robustness}
To see how robust a given classifier $f(\cdot)$ is against single-word perturbation, we take the training set $\Dtrain$, and find the subset $\Dtrain^+$ where $f$ can correctly predict the label. We want to find out whether each word $w_j \in V$ can flip the prediction of each example sentence $(\bx_i, y_i) \in \Dtrain^+$. Therefore, we define a matrix  $A\in [0,1]^{|\Dtrain^+|\times |V|}$ indicating whether a word can successfully flip the classifier's prediction of a sentence, specifically

\begingroup
\small%
\begin{equation}
A_{i,j} = 
\begin{cases}
1 & \text{if } \exists \bx' \in S(\bx_i, w_j) \text{ s.t. } f(\bx') \neq y_i\\
0 & \text{otherwise}
\end{cases}
\end{equation}
\endgroup

\noindent where $S(\bx, w)$ is a set of sentences constructed by replacing one word in $\bx$ with $w$. Note that $A$ only indicates that the prediction was flipped, and does not take into account similarity or fluency.

With this matrix, we can find the \textit{single-word flip capability} for each word in the classifier's vocabulary, denoted as \ap and defined as the percentage of sentences whose prediction can be successfully flipped using that word. It can be computed as 

\begingroup
\small%
\[\kappa(w_j) = \frac{1}{|\Dtrain^+|}\sum_{i} A_{i,j}\]
\endgroup

Second, we can find the \textit{robustness against single-word  perturbations}, for a text classifier, denoted as \spa and defined as the percentage of sentence-word pairs in the Cartesian product of the dataset and the vocabulary where the classifier prediction cannot be successfully flipped. It can be computed as 

\begingroup
\small%
\[\rho(f) = \frac{1}{|\Dtrain^+|\cdot|V|}\sum_{i,j} (1-A_{i,j})\]
\endgroup


\subsection{Definitions and Theorem}
We then give formal definitions and a theorem.
\begin{definition}
The single-word flip capability of word $w$ on a classifier $f$ is 

\begingroup
\small%
\begin{equation*}
\kappa_{f}(w) = 
\frac{|\{(\bx, y) {\in \Dtrain^{+}}| \exists \bx'\in S(\bx, w) \text{ s.t. } f(\bx')\neq y\}|}{|\Dtrain^{+}|}
\end{equation*}
\endgroup

\noindent where $\Dtrain^{+}=\{(\bx, y) \in \Dtrain | f(\bx)=y\}$ is a subset of $\Dtrain$ that is correctly classified by $f$, and
$S(\bx, w)$ is a set of sentences constructed by replacing one word in $\bx$ with $w$. We omit $f$ and use notation $\kappa(w)$ in the rest of the paper.
\end{definition}

\begin{definition}
The robustness against single-word perturbation is

\begingroup
\small%
\begin{equation*}
\rho(f) = \frac{|\{((\bx, y), w)\in \Dtrain^*| \forall \bx' \in S(\bx, w): f(\bx') = y\}|}{|\Dtrain^*|}
\end{equation*}
\endgroup

\noindent where $\Dtrain^* = \Dtrain^{+} \times V$ is the Cartesian product of $\Dtrain^+$ and vocabulary $V$. 
\end{definition}
$\rho(f)$ can be interpreted as the accuracy of $f$ on $\Dtrain^*$, where $f$ is considered correct on $((\bx, y), w)$ if all sentences in $S(\bx, w)$ are predicted as $y$.

\begin{theorem} 
\ap and \spa have the following relation:
\label{chap4-thm1}

\begingroup
\small%
\begin{equation*}
{
\rho(f) = 1 - \frac{1}{|V|} \sum_{w\in V} \kappa_f(w)
}
\end{equation*}

\textit{Proof:}
\begin{align*}
    \rho(f) &= \frac{\sum_{i,j} 1 - A_{i,j}}{|\Dtrain^{+}| \times |V|}\\  
    &= 1 - \frac{1}{|V|} \sum_{j=1}^{|V|} \frac{\sum_{i=1}^{|\Dtrain^{+}|} A_{i,j}}{|\Dtrain^{+}|}\\
    &= 1 - \frac{1}{|V|} \sum_{j=1}^{|V|} \kappa(w_j)
\end{align*}
\endgroup
\end{theorem}

\section{Efficient Estimation of the Metrics}\label{sec:alg}
Directly computing \ap and \spa metrics is time-consuming, because each word in the vocabulary needs to be placed in all sentences at all positions and then verified on the classifier. To overcome this issue, we instead estimate the lower bound of \ap (and thus the upper bound of \spa). In this section, we first give an overview of the efficient upper bound algorithm (\alg), then detail our first-order approximation. Figure~\ref{fig:euba} depicts the algorithm and the pseudocode is presented in Appendix B.

\subsection{Overview}
To find the lower bound of $\kappa(w)$, we want to find as many $(\bx, y) \in D^{+}$ such that $w$ can successfully flip the prediction. Since we want to compute \ap for all words, we can convert it to the symmetric task of finding as many $w \in V$ as we can for each $(\mathbf{x},y) \in D^+$ such that the sentence can be successfully flipped by the word. This conversion enables a more efficient algorithm.

For each example $(\bx, y)$, where $\bx = x_1, \ldots, x_l$, each word can be substituted with any $w\in V$, leading to a total of $l\times|V|$ substitutions. For a substitution $x_k \rightarrow w$, we compute 

\vspace{-1em}
\begingroup
\small%
\begin{equation}
{\textstyle
    u^{(k)}_{w} \approx \log f(y|x_1, \ldots, x_{k-1}, w, x_{k+1},\ldots, x_l)}
\end{equation}
\endgroup

\noindent where $f(y|\cdot)$ is the classifier's probability of predicting $y$. We will show the computation of $u^{(k)}_{w}$ in Section~\ref{sec:approx}. We assume substitutions with lower $u^{(k)}_{w}$ are more likely to flip predictions and verify them in two phases. 

\noindent\textbf{Phase 1:} We sort all substitutions in ascending order by $u^{(k)}_{w}$, and verify them on the classifier. We stop the verification after $m$ consecutive unsuccessful attempts and assume all remaining substitutions are unsuccessful, where $m$ is a hyper-parameter. 

\noindent\textbf{Phase 2:} If a word can successfully flip many other sentences in the class $y$, it is more likely to succeed on $\bx$. Therefore, we keep track of how many successful flips each word can trigger on each category. We sort all words in descending order by the number of their successful flips, and verify them. {For each word $w$, we only verify the position where it is most likely to succeed (i.e., $\arg\min_k u^{(k)}_{w}$)}. Similarly, phase 2 stops after $m$ consecutive unsuccessful attempts. 

By using $u^{(k)}_{w}$, we can skip a lot of substitutions that are unlikely to succeed, improving efficiency. The hyper-parameter $m$ controls the trade-off between efficiency and the gap between the lower bound and the exact \ap.
When setting $m\rightarrow\infty$, \alg can find the exact \ap and \spa. In Section~\ref{sec:ablation}, we show that efficiency can be improved a lot if a relatively small $m$ is used, at the small cost of neglecting some successful flips. We also compare \alg with two alternative designs. 

\subsection{First-order Approximation}\label{sec:approx}
We then show how to efficiently compute $u^{(i)}_{w}$ using first-order approximation. We construct the following structure $S(\bx, \mask)=$
\[
\begin{small}
\begin{Bmatrix}
    \bs_1:& \mask, x_2, x_3, \ldots, x_l; \\
    \bs_2:& x_1, \mask, x_3, \ldots, x_l; \\
    & \ldots\\
    \bs_l:& x_1, x_2, \ldots, x_{l-1}, \mask; \\
\end{Bmatrix}.
\end{small}
\]
In a typical classifier, the input sentence will be converted to a sequence of embeddings noted $\be_{\cdot}$; thus we convert $S(\bx, \mask)$ to 
\[
\begin{small}    
\begin{Bmatrix}
    E_{\bs_1}:& \be_\mask, \be_{x_2}, \be_{x_3}, \ldots, \be_{x_l}; \\
    E_{\bs_2}:& \be_{x_1}, \be_{\mask}, \be_{x_3}, \ldots, \be_{x_l}; \\
    & \ldots\\
    E_{\bs_l}:& \be_{x_1}, \be_{x_2}, \ldots, \be_{x_{l-1}}, \be_\mask; \\
\end{Bmatrix}.
\end{small}
\]

\begin{figure*}[!t] 
    \centering
    \includegraphics[width=0.99\textwidth]{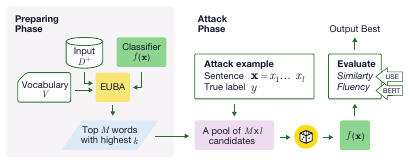}
    \caption{\attack pipeline. High-flip-capacity words are used to conduct low-cost single-word adversarial attacks.}
    \label{fig:attack}
\end{figure*}

We input $\bs_k$ into the classifier, then compute the gradient of $\log f(y|\bs_k)$ with respect to $\be_\mask$ at the $k$-th position. We then approximate the log probability by substituting $\be_\mask$ with $\be_{w}$. The log probability change can be approximated by the inner product of $\nabla_{\be_{\mask}} \log f(y|\bs_k)$ and $\be_w - \be_\mask$. Formally,

\begingroup
\small%
\begin{align}
u_{w}^{(k)} =& \langle \nabla_{\be_{\mask}} \log f(y|\bs_k), \be_{w} - \be_{\mask} \rangle \nonumber \\
    &+\log f(y|\bs_k)
\label{eq:approx}
\end{align}
\endgroup

\noindent $f(y|\bs_k)$ and $\nabla_{\be_{\mask}} \log f(y|\bs_k)$ needs to be computed only once for each masked sentence $\bs_k$, and does not have to repeat for every word-and-sentence combo. $\be_{w}$ and $\be_\mask$ are fixed vectors. Eq.~(\ref{eq:approx}) is an inner product, and thus very efficient to compute. 

\section{Single-word Perturbation Attack}\label{sec:attack}
For attackers who try to find fluent and semantically similar adversarial examples, words with high \ap provide the potential for low-cost attacks. We propose \attack, which first uses our algorithm, \alg to estimate \ap, then use the top $M$ words with highest \ap to craft an attack. Figure~\ref{fig:attack} shows the attack pipeline.

To conduct an attack on a sentence $\bx$ of length $l$, we try to put these words at all possible positions to create a pool of $l\times M$ candidate sentences. Then, we draw samples from the pool and verify them on the classifier. We stop when we either exhaust the pool or find $k$ candidates that change the prediction. To ensure similarity and fluency, we employ Universal Sentence Encoder (USE)~\cite{cer2018universal} and BERT language models. For each $\bx'$ that can change the prediction, we compute a joint criteria score 

\vspace{-1em}
\begingroup
\small%
\begin{equation}\label{eq:attack}
    \alpha \cos(H(\bx'), H(\bx)) - \beta \frac{\ppl(\bx')}{\ppl(\bx)}
\end{equation}
\endgroup

\noindent where $H(\cdot)$ is the USE sentence embedding, and $\ppl(\cdot)$ is the perplexity of a sentence measured by BERT. We then pick the sentence with the highest score as output. We set $M=50, k=50, \alpha=3, \beta=20$ in our experiments. 

There are two differences between \attack and conventional black-box adversarial attacks. (1)~\attack substitutes only one word with one of the $M=50$ words, whereas conventional methods can change multiple words and can use any word. As a result, \attack is much more efficient. (2)~Estimating \ap using \alg requires computing the gradient of the classifier, while black-box adversarial attack methods do not. Even so, our attack method can be applied to real-world scenarios -- for example, an insider attack where someone with access to the classifier calculates the flip capacity and leaks high-capacity words to other malicious users, who use them to conduct attacks.

\section{Single-word Perturbation Defense}\label{sec:defense}
We present a data augmentation strategy, \defense, to improve the robustness of the classifier in a single-word perturbation scenario.
Specifically, we design three augmentations. 

\noindent\textbf{Random augmentation} randomly picks one word in a sentence, then replaces it with another random word in $V$. Since $95\%$ words have at least 6.5\% \ap according to our experimental results, even pure randomness can sometimes alter the classifier's prediction. 

\noindent\textbf{Gradient-based augmentation} uses gradient information to find a single-word substitution that is more likely to cause misclassification. We approximate the log probability of correct prediction after substituting $x_i$ with $w$ as 

\begingroup
\small%
\begin{equation}
{\textstyle
v_{w}^{(i)} = \langle \nabla_{\be_{x_i}} \log f(y|\bx), \be_{w} - \be_{x_i} \rangle +\log f(y|\bx)}
\label{eq:approx2}
\end{equation}
\endgroup

Then we apply the substitution with the minimum $v_{w}^{(i)}$. Eq.~(\ref{eq:approx2}) is more efficient because it only needs one forward and backward pass on $f(y|\bx)$ to compute $\nabla_{\be_{x_i}} \log f(y|\bx)$, comparing to $l$ forward and backward passes on $\nabla_\mask \log f(y|\bs_i)$ in Eq.~(\ref{eq:approx}). Therefore it is more suitable for data augmentation, which usually involves large-scale training data. 

Some attack methods use a vocabulary different from $V$. They can substitute $x_i$ with a word $t\not\in V$. Neither gradient-based nor random augmentation can defend against adversarial examples caused by $t$. \noindent\textbf{Special word augmentation} is designed to address this issue. For each class $y$, we find a set of words that occurs much more frequently in other classes $y'$, 
formally

\vspace{-1em}
\begingroup
\small%
\begin{align}
    T(y) = \{t | &\max_{y'\neq y} \log \text{freq}_D(t, y') \nonumber\\
                & - \log \text{freq}_D(t, y) > 1\}
\end{align}
\endgroup

\noindent where $\text{freq}_D(t, y)$ is the frequency of the word $t$ in all the training examples with the label $y$. To augment an example $(\bx, y)$, we randomly sample a position in $\bx$ and replace it with a random word $t\in T(y)$.

In each training iteration, we apply gradient-based augmentation on half of the batch. For the other half, we randomly choose from original training data, random augmentation, or special word augmentation.

\section{Experiments}\label{sec:exp}

In this section, we conduct comprehensive experiments to support the following claims:
    \ap and \spa metrics are necessary in showing the classifier robustness against single-word perturbation   (Section~\ref{sec:exp_metric});
    \attack is as effective as conventional adversarial attacks that change multiple words  (Section~\ref{sec:exp_att});
    \defense can improve any classifier's robustness in both single-word and multiple-word perturbation setups (Section~\ref{sec:exp_def});
    \alg is well-designed and configured to achieve a tighter bound than alternative designs (Section~\ref{sec:ablation}).

\begin{table}[htb]
\small
\centering
\caption{Dataset details. \textbf{\#C} is the number of classes, \textbf{Len} is the average number of words in a sentence.}\label{tab:dataset}
\vspace{-1em}
\begin{tabular}{lccp{4.5cm}}
\toprule
\textbf{Name} & \textbf{\#C} & \textbf{Len} & \textbf{Description}\\
\midrule
AG  & 4 & 43 & News topic classification.    \\
MR & 2 & 32 & Movie review dataset by \citet{pang2005seeing}.\\
SST2 & 2 & 20 & Binary Sentiment Treebank \citep{wang2018glue}.\\
HATE & 2 & 23 & Hate speech detection dataset by \citet{kurita20acl}.\\
\bottomrule
\end{tabular}
\end{table}

\subsection{Setup}\label{sec:setup}
\noindent \textbf{Datasets.} We conduct our experiments on 4 datasets: 1) \textbf{AG} -- news topic classification, 2) \textbf{MR} -- movie review sentiment classification~\citep{pang2005seeing}, 3) SST2 -- Binary Sentiment Treebank \citep{wang2018glue}, and 4) \textbf{Hate} speech detection dataset \citep{kurita20acl}. Dataset details are shown in Table~\ref{tab:dataset}.

\noindent \textbf{Classifiers.} We evaluate our method on two classifiers, the BERT-base classifier~\cite{Devlin2019BERT} and the distilBERT-base classifier~\cite{sanh2019distilbert}. {We select BERT models, which are masked language models, because they are trained to predict the masked token}. The \textit{vanilla classifiers} are trained with the full training set for 20k batches with batch size 32 using the AdamW~\cite{loshchilov2017adamw} optimizer and learning rate 2e-5.
We also include results for the RoBERTa-base classifier~\cite{liu2019roberta} in Appendix E.3.

\noindent \textbf{Metrics.} We use several metrics to show the quality and robustness of a classifier. 
\begin{itemize}[leftmargin=*]
    \itemsep-0.1em
    \item (\textbf{CAcc}$\uparrow$): clean accuracy, the accuracy of the classifier measured on the original test set.
    \item \spa: quantifies classifier robustness in a single-word perturbation scenario.
    \begin{itemize}
        \itemsep0em
        \item \textbf{\spa}($\uparrow$) is measured on 1000 examples sampled from the training set. 
        \item \textbf{\spat}($\uparrow$) is measured on the test set. 
    \end{itemize}  
    \item (\textbf{ASR} $\downarrow$): attack success rate that shows how robust the classifier is against an adversarial attack method. This is defined as 
    \[
    {\textstyle
    ASR = \frac{\text{\# successful attacks}}{\text{\# correct prediction in testset}}
    }
    \] 
    A lower ASR means the classifier is more robust. We consider an attack to be successful if the similarity measured by the cosine of USE embeddings is greater than 0.8.
    \item (\textbf{ASR1}$\downarrow$): Since we are interested in single-word perturbations, we constrain other attack methods and use adversarial examples from them that have only one word perturbed and evaluate ASR. 
\end{itemize}
Note that ASR also shows the efficacy of attack methods. A higher ASR means the attack method is more effective. We use ASR$(\uparrow)$, and similarly ASR1$(\uparrow)$, in the context of comparing attacks. 

\noindent\textbf{Human evaluation.} We collect human annotation to further verify the similarity and fluency of adversarial sentences. We ask human annotators to rate sentence \textbf{similarity} and sentence \textbf{fluency} in 5-likert. In addition, we ask humans to check if the adversarial sentence \textbf{preserves the original label}. See Appendix E.2 for detailed settings.

\noindent \textbf{Adversarial attack SOTA.}  We compare \attack against three state-of-the-art methods, namely:  \textbf{TextFooler}~\cite{Jin2019textfooler}, \textbf{BAE}~\cite{garg2020bae}, and \textbf{CLARE}~\cite{li2021CLARE}. 
When reporting ASR, any adversarial example generated with one (or more) word changes is considered a successful attack. 
While in ASR1, we consider the subset of adversarial sentences with only a single-word change as a successful attack.


\noindent \textbf{Data augmentation and adversarial training. } We compare \defense with several baselines. 
\begin{itemize}[topsep=1pt, partopsep=1pt, leftmargin=12pt, itemsep=-2pt]
\item \textbf{Rand}: We conduct random perturbations to augment training data in all steps of training. 
\item \textbf{A2T}~\cite{yoo2021a2t} is an efficient gradient-based adversarial attack method designed for adversarial training. 
\end{itemize}
For Rand and \defense, we tune the classifier with another 20k batches. For A2T, we tune the classifier with 3 additional epochs as recommended by the authors.

\begin{figure}[tb]
    \centering
    \vspace{-1em}
    \includegraphics[width=0.95\columnwidth]{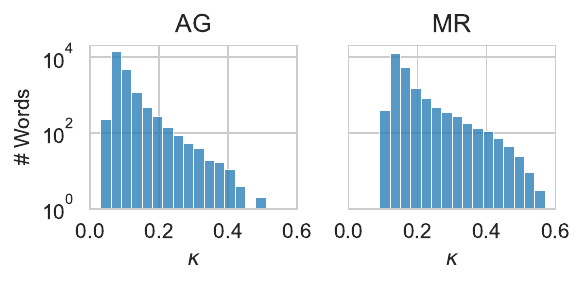}
    \vspace{-1em}
    \caption{Histogram of words at different \ap. Note that the y-axis is in log scale.}
    \vspace{-1em}
    \label{fig:word_hist}
\end{figure}

\subsection{Necessity of \ap and \spa metrics}\label{sec:exp_metric}



\begin{figure*}[tb]
    \centering
    \itemsep0em
    \includegraphics[width=0.45\textwidth]{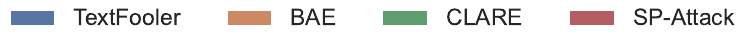}\\
    \includegraphics[width=0.4\textwidth]{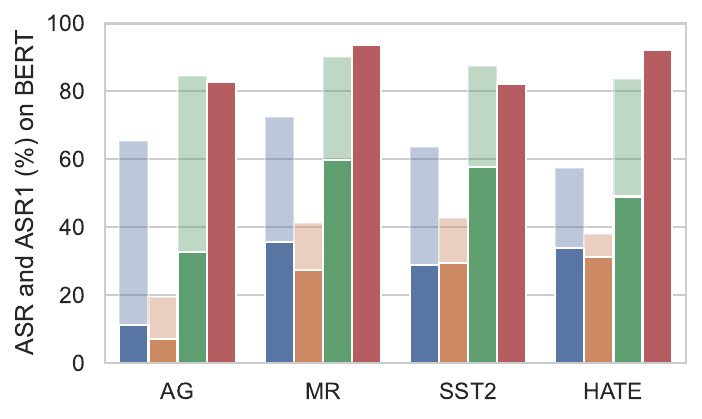}
    \includegraphics[width=0.4\textwidth]{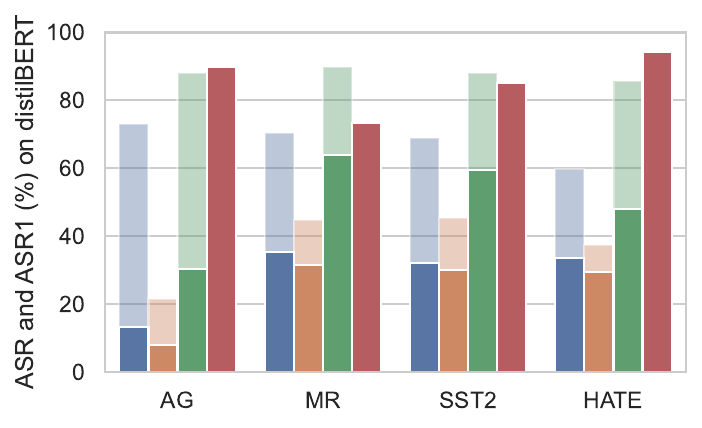}
    \caption{ASR($\uparrow$) and ASR1($\uparrow$) on vanilla BERT-base (left) and distilBERT-base (right). The translucent (taller) bars represent ASR, while the solid (shorter) bars represent ASR1. For \attack, ASR and ASR1 are the same.}
    \label{fig:asr_vanilla}
\end{figure*}

We measure the exact \ap using brute force for the BERT-base classifier on AG and MR. The measurement takes 197 and 115 GPU hours respectively on a single Nvidia V100 GPU. Figure~\ref{fig:word_hist} shows the histogram of \ap. We find that 95\% of words in the vocabulary can successfully attack at least 6.5\% and 12.4\% of examples, while the top 0.1\% of words can change as much as 38.7\% and 49.4\% examples on the two datasets respectively. This shows that classifiers are extremely vulnerable to single-word perturbations. The distributions of \ap for the two tasks are different, showing \ap is task-dependent. Words have lower average \ap on AG than on MR. We compute the $\rho(f)$ as $90.9$ and $83.8$ for the two datasets respectively. 
See additional analysis of words with high \ap in Appendix D.


We highlight the necessity of \ap and \spa metrics. \ap can show the flip capability of each individual word. It precisely reveals all vulnerabilities of the classifier under single-word perturbation. \spa is derived from \ap and factors in the number of different single-word perturbations that can successfully attack each sentence. It quantifies classifier robustness to single-word perturbation scenarios more effectively than ASR, which only shows whether each sentence could be successfully attacked.

\begin{table}[tb]
    \centering
    \caption{Similarity scores, fluency scores, and the percentage of sentences which preserve the original label by human annotators.}
    \resizebox{\columnwidth}{!}{%
    \begin{tabular}{lcccc}
        \toprule
        \textbf{Dataset} & \textbf{Attack} & \textbf{Similarity} & \textbf{Fluency} &  \textbf{Preserve Label} \\
        \midrule
        AG & TextFooler     & 3.40 & 3.41 & 80\% \\
           & Clare          & 3.37 & 3.33 & 82\% \\
           & SP-Attack      & 3.47 & 3.39 & 87\% \\
        \midrule
        MR & TextFooler     & 3.93 & 3.47 & 69\% \\
           & Clare          & 3.96 & 3.41 & 83\% \\
           & SP-Attack      & 4.02 & 3.48 & 93\% \\
        \bottomrule
    \end{tabular}}
    \label{tab:human-anno}
\end{table}

\begin{table*}[tb]
\centering
\caption{CAcc($\uparrow$), \spa($\uparrow$) and ASR($\downarrow$) of \attack on vanilla and improved classifiers. NA denotes the vanilla classifier. \defense is abbreviated as \deshort.}\label{tab:spar}\vspace{-1em}
\begingroup
\resizebox{\linewidth}{!}{%
\begin{tabular}{clcccccccccccccccc}
\toprule
 &      & \multicolumn{4}{c}{\textbf{AG}} & \multicolumn{4}{c}{\textbf{MR}} & \multicolumn{4}{c}{\textbf{SST2}} & \multicolumn{4}{c}{\textbf{HATE}} \\
 \cmidrule(lr){3-6}\cmidrule(lr){7-10}\cmidrule(lr){11-14}\cmidrule(lr){15-18}
 & \textbf{Defense} & \textbf{CAcc}  & \textbf{\spa}  & \textbf{\spat} & \textbf{ASR}  & \textbf{CAcc}  & \textbf{\spa}  & \textbf{\spat} & \textbf{ASR}  & \textbf{CAcc}   & \textbf{\spa}   & \textbf{\spat} & \textbf{ASR}  & \textbf{CAcc}   & \textbf{\spa}   & \textbf{\spat} & \textbf{ASR}  \\\midrule
\multirow{4}{*}{\rotatebox[origin=c]{90}{\textbf{BERT}}}
&NA       & 93.7 & 91.5 & 90.1 & 82.7 & 87.7 & 86.4 & 77.2 & 93.5 & \textbf{80.6} & 74.7 & 76.6 & 82.1 & \textbf{94.5} & 72.6 & 71.4 & 92.1 \\\cmidrule(lr){2-2}
&Rand     & 92.0 & 96.3 & 93.1 & 66.2 & 87.5 & 98.2 & 80.7 & 84.2 & 79.8 & 82.1 & 79.7 & 77.9 & 94.3 & 86.3 & 83.2 & 89.2 \\
&A2T      & 93.4 & 96.2 & 92.9 & 62.7 & \textbf{88.3} & 90.4 & 76.2 & 69.3 & 80.4 & 76.5 & 74.9 & 69.5 & 93.4 & 87.1 & 82.1 & 87.3 \\
&\deshort & \textbf{94.3} & \textbf{97.4} & \textbf{94.1} & \textbf{33.7} & 87.5 & \textbf{99.8} & \textbf{83.1} & \textbf{61.8} & 79.3 & \textbf{90.0} & \textbf{80.5} & \textbf{68.8} & 93.7 & \textbf{96.6} & \textbf{92.5} & \textbf{64.4} \\
\midrule
\multirow{4}{*}{\rotatebox[origin=c]{90}{\textbf{distilBERT}}}
&NA       & 94.0 & 91.7 & 91.6 & 89.9 & 85.9 & 85.6 & 72.7 & \textbf{73.2} & \textbf{78.7} & 75.8 & 74.3 & 85.1 & \textbf{93.3} & 73.8 & 72.9 & 94.2 \\\cmidrule(lr){2-2}
&Rand     & 94.2 & 95.4 & 92.0 & 69.5 & \textbf{86.0} & 96.5 & 78.1 & 92.2 & 78.0 & 81.1 & 78.3 & 87.2 & 93.1 & 85.6 & 82.6 & 91.3 \\
&A2T      & 94.0 & 95.3 & 92.6 & 64.8 & 85.0 & 85.1 & 73.4 & 88.8 & 78.3 & 77.7 & 76.1 & 80.5 & 93.2 & 86.1 & 82.0 & 89.5 \\
&\deshort & \textbf{94.3} & \textbf{97.6} & \textbf{93.6} & \textbf{43.1} & 85.1 & \textbf{99.8} & \textbf{79.6} & 73.4 & 78.2 & \textbf{88.3} & \textbf{80.2} & \textbf{73.0} & 92.0 & \textbf{96.9} & \textbf{91.6} & \textbf{68.4} \\
\bottomrule
\end{tabular}}
\endgroup
\vspace{-2ex}
\end{table*}

\begin{figure}[htb!]
    \centering
    \includegraphics[width=0.95\columnwidth]{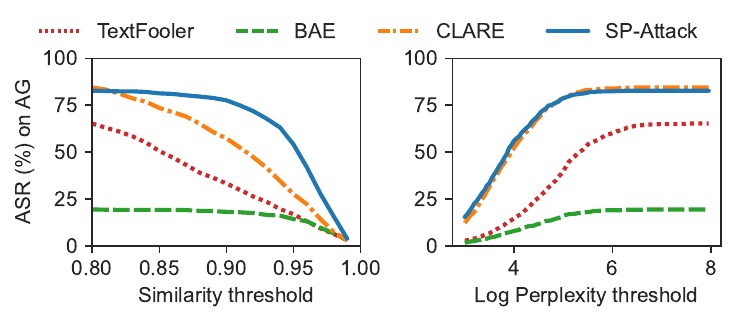}
    \includegraphics[width=0.95\columnwidth]{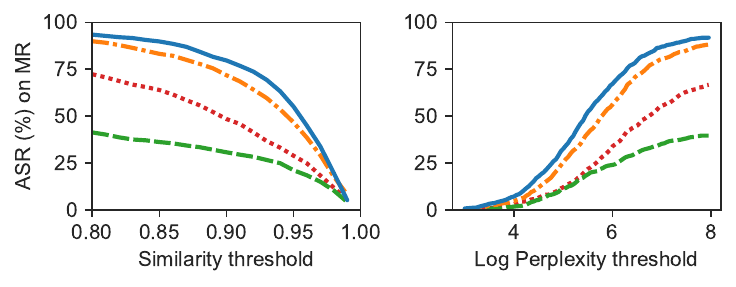}
    \caption{Comparing ASR($\uparrow$) under different similarity and perplexity thresholds on AG and MR datasets with the BERT-base classifier. }
    \label{fig:asr_threshold}
    \vspace{-3ex}
\end{figure}
\begin{figure*}[tb]
    \centering
    \includegraphics[width=.9\textwidth]{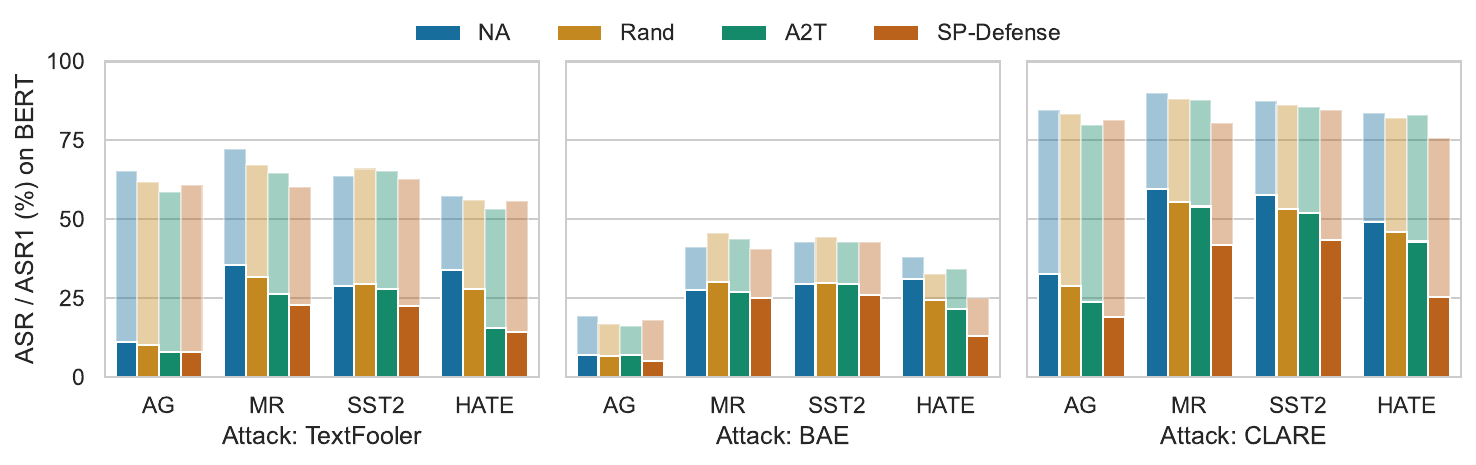}\\
    \vspace{-1ex}
    \includegraphics[width=.9\textwidth]{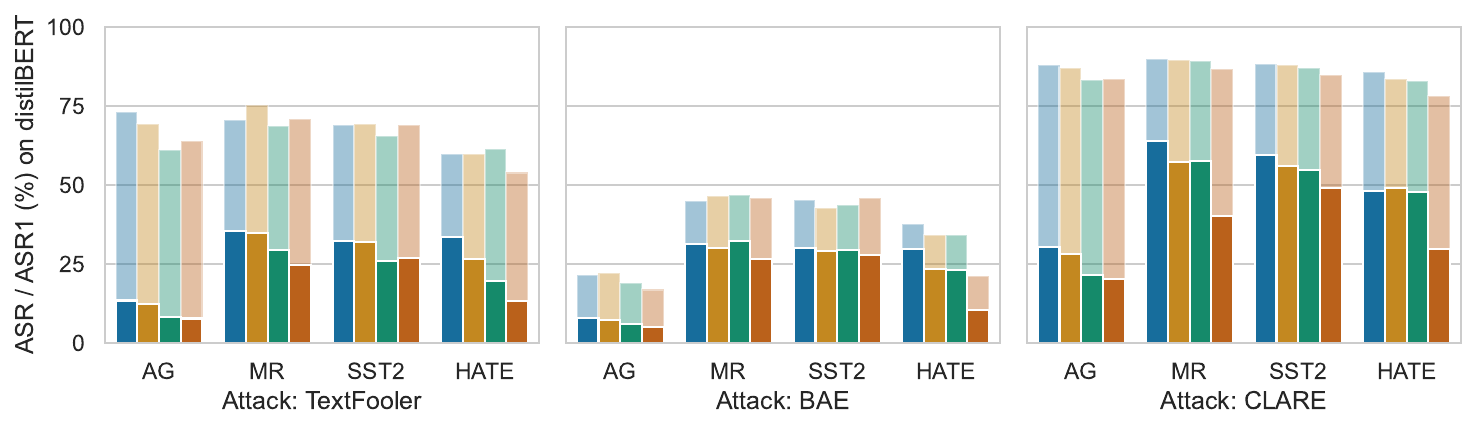}
    \vspace{-2ex}
    \caption{ASR($\downarrow$) and ASR1($\downarrow$) of TextFooler, BAE and CLARE on vanilla and improved BERT-base (top) and distilBERT-base (bottom) classifiers. NA denotes the vanilla classifier.}
    \vspace{-3ex}
    \label{fig:improved_asr}
\end{figure*}

\subsection{\attack performance}\label{sec:exp_att}
Figure~\ref{fig:asr_vanilla} shows the ASR and ASR1 for all datasets, two different classifiers (left and right) and multiple attack methods. We make a few noteworthy observations. \attack is the most effective method for finding adversarial examples based on single-word perturbations. ASR1's for \attack are at least 73\%, and  over 80\% in 7 out of 8 cases, demonstrating the significant vulnerability of these classifiers. A second observation is that other methods, even those with a notable number of single word perturbation-based adversarial examples, have significantly lower ASR1's (where only single-word adversarial examples are included) than ASR's (where multi-word adversarial examples are included). For TextFooler across the 8 experiments, the difference between ASR and ASR1 is 38.3\%. 
An obvious next question we examine is whether multi-word substitutions are essential for achieving higher attack success rates. Our experiments also show that \attack achieves comparable ASR with single-word perturbations, and even beats the best multi-word substitution in 4 out of 8 cases.

Single-word perturbation may cause poor fluency and/or low semantic similarity. To show \attack is comparable with multi-word attacks in terms of fluency and similarity, we plot the ASR under different similarity and perplexity thresholds. We measure similarity using USE and perplexity using a BERT language model fine-tuned on the dataset. Figure~\ref{fig:asr_threshold} shows the results on AG and MR datasets. It shows that, at the same threshold, \attack achieves comparable or better ASR than those attack baselines. Table~\ref{tab:human-anno} shows the human evaluation. The similarity and fluency scores are similar for all methods, but \attack outperforms baselines by 5\% and 10\% in preserving the original label on AG and MR datasets respectively. We show inter-annotator agreement in Appendix E.2.

\subsection{\defense Performance}\label{sec:exp_def}
\noindent \textbf{Can our \defense overcome our attack?} We measure the accuracy and robustness of the vanilla and improved classifiers. The complete results are available in Appendix E.3.

Table~\ref{tab:spar} shows the CAcc, \spa, \spat, and ASR of \attack. We observe that all data augmentation or adversarial training methods have small impacts on CAcc; in most cases, the decrease is less than 1\%. However, \spa and \spat differ a lot, showing that classifiers with similar accuracy can be very different in terms of robustness to single-word attacks (both in training and testing). We found that \defense outperforms Rand and A2T on \spat in all cases. This shows that \defense can effectively improve a classifier's robustness in a single-word perturbation scenario. Averaged over 4 datasets, \defense achieves 14.6\% and 13.9\% increases on \spa; 8.7\% and 8.4\% increases on \spat; and 30.4\% and 21.2\% decreases on ASR of \attack on BERT and distilBERT classifiers respectively.

\noindent \textbf{Can our \defense overcome other SOTA attacks constrained to single-word adversarial examples?} Figure~\ref{fig:improved_asr} shows the ASR and ASR1 on vanilla and improved classifiers for different SOTA attacks (left to right). The ASR1 decreases by a large margin after the application of \defense, which is consistent with the improvement on \spat. The ASR also decreases in 20 out of 24 cases, showing the improvement of classifier robustness against a conventional multi-word adversarial attack setup.

\subsection{Justifying Design Decisions of \alg}\label{sec:ablation}

We show the gap between the \spa upper bound and its true value when setting different early stop criteria. We set $m=\{128, 256, 512, 1024, 2048\}$. We also changed phase 1 in \alg to form two alternative designs:
\begin{itemize}
\item \alg(top1): We use the same gradient-based approximation. But for each word, we only verify the top position, i.e. $\arg\min_i u^{(i)}_w$.
\item \alg(w/o mask): We do not replace each position with masks. Instead, we use the original sentence to directly compute a first-order approximation like Eq.~(\ref{eq:approx2}). 
\end{itemize}

\begin{figure}[tb]
    \centering
    \includegraphics[width=\columnwidth]{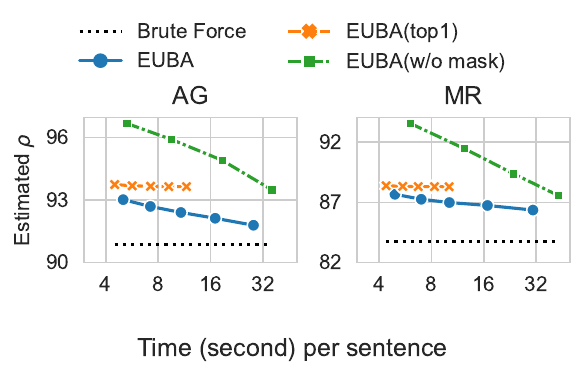}
    \caption{Comparing the estimated \spa with respect to time. For each method, the early stop criteria is 128, 256, 512, 1024, and 2048 (from left to right on each curve). \alg(w/o mask) 2048 is not shown in the figure because of long time consumption. The Brute Force method takes 720s and 415s per sentence on AG and MR respectively.}
    \label{fig:ara_time}
    \vspace{-3ex}
\end{figure}

Figure~\ref{fig:ara_time} shows the estimated \spa with respect to average time consumption for each example in $D^+$. We observe that our proposed design is the closest to brute force. \alg outperforms alternative methods given the same time budget, and is very close to the true \spa.

\vspace{-0.8ex}
\section{Related Work}
\vspace{-0.8ex}


The widespread deployment of text classifiers has exposed several vulnerabilities. {Backdoors can be injected during training that can later be used to manipulate the classifier output}~\cite{li2021backdoor, zhang2021red, jia2022badencoder, yan2023bite, you2023betteradversaries, du2024backdoor}. The use cases of backdoor attacks are limited because they involve interfering with the training. Universal adversarial triggers can be found for classifiers~\cite{wallace2019universal}, wherein triggers, usually involving multiple-word perturbations, change the classifier prediction for most sentences. {Adversarial attack methods}~\cite{Jin2019textfooler, garg2020bae, li2020bertattack, li2021CLARE, zhan2024rethinking, zhan2023similarizing, zhang2023semantics, li2023adversarial, ye2022texthoaxer} change a classifier's prediction by perturbing a few words or characters in the sentence. They maintain high semantic similarity and fluency, but are computationally expensive. \attack attacks the classifier by trying just a few single-word perturbations, making it an efficient attack that maintains the high similarity and effective fluency of conventional adversarial attacks. 

To defend against attacks, adversarial training can be applied. Adversarial training is often used to help the classifier resist the types of attacks described above. A2T~\cite{yoo2021a2t} is an efficient attack designed specifically for adversarial training. There are other defense methods including perplexity filtering~\cite{qi2020onion}, and synonym substitution~\cite{wang2021synonymencoding}, which increase the inference time complexity of the classifier. Our \defense does not add overhead to inference and can effectively defend against adversarial attacks.

\section{Conclusion}
In this paper, we comprehensively analyze a restricted adversarial attack setup with single-word perturbation. We define two useful metrics -- \spa, the quantification of a text classifier's robustness against single-word perturbation attacks, and \ap, the adversarial capability of a word -- and show that these metrics are useful to measure classifier robustness. Furthermore, we propose \attack, a single-word perturbation attack which achieves a comparable or better attack success rate than existing and more complicated attack methods on vanilla classifiers, and significantly reduces the effort necessary to construct adversarial examples. Finally, we propose \defense to improve classifier robustness in a single-word perturbation scenario, and show that it also improves robustness against multi-word perturbation attacks. 

\section*{Acknowledgement}
Prof. Cuesta-Infante was supported by R\&D project TED2021-129162B-C22, funded by MICIU/AEI/10.13039/501100011033/ and the European Union NextGenerationEU/ PRTR, and R\&D project PID2021-128362OB-I00, funded by MICIU/AEI/10.13039/501100011033/ and FEDER/UE.

\bibliography{ref}

\begin{thebibliography}{30}
\expandafter\ifx\csname natexlab\endcsname\relax\def\natexlab#1{#1}\fi

\bibitem[{Cer et~al.(2018)Cer, Yang, Kong, Hua, Limtiaco, St.~John, Constant,
  Guajardo-Cespedes, Yuan, Tar, Strope, and Kurzweil}]{cer2018universal}
Daniel Cer, Yinfei Yang, Sheng-yi Kong, Nan Hua, Nicole Limtiaco, Rhomni
  St.~John, Noah Constant, Mario Guajardo-Cespedes, Steve Yuan, Chris Tar,
  Brian Strope, and Ray Kurzweil. 2018.
\newblock Universal sentence encoder for english.
\newblock In \emph{Proceedings of the Conference on Empirical Methods in
  Natural Language Processing: System Demonstrations}.

\bibitem[{Devlin et~al.(2019)Devlin, Chang, Lee, and
  Toutanova}]{Devlin2019BERT}
Jacob Devlin, Ming-Wei Chang, Kenton Lee, and Kristina Toutanova. 2019.
\newblock Bert: Pre-training of deep bidirectional transformers for language
  understanding.
\newblock In \emph{Proceedings of the Annual Conference of the North American
  Chapter of the Association for Computational Linguistics}.

\bibitem[{Du et~al.(2024)Du, Ju, Ren, Li, and Liu}]{du2024backdoor}
Wei Du, Tianjie Ju, Ge~Ren, GaoLei Li, and Gongshen Liu. 2024.
\newblock \href {https://aclanthology.org/2024.lrec-main.186} {Backdoor {NLP}
  models via {AI}-generated text}.
\newblock In \emph{Proceedings of the 2024 Joint International Conference on
  Computational Linguistics, Language Resources and Evaluation (LREC-COLING
  2024)}, pages 2067--2079, Torino, Italia. ELRA and ICCL.

\bibitem[{Garg and Ramakrishnan(2020)}]{garg2020bae}
Siddhant Garg and Goutham Ramakrishnan. 2020.
\newblock Bae: Bert-based adversarial examples for text classification.
\newblock In \emph{Proceedings of the Conference on Empirical Methods in
  Natural Language Processing and the International Joint Conference on Natural
  Language Processing}.

\bibitem[{Jia et~al.(2022)Jia, Liu, and Gong}]{jia2022badencoder}
Jinyuan Jia, Yupei Liu, and Neil~Zhenqiang Gong. 2022.
\newblock \href {https://doi.org/10.1109/SP46214.2022.9833644} {Badencoder:
  Backdoor attacks to pre-trained encoders in self-supervised learning}.
\newblock In \emph{2022 IEEE Symposium on Security and Privacy (SP)}, pages
  2043--2059.

\bibitem[{Jin et~al.(2020)Jin, Jin, Zhou, and Szolovits}]{Jin2019textfooler}
Di~Jin, Zhijing Jin, Joey~Tianyi Zhou, and Peter Szolovits. 2020.
\newblock Is bert really robust? natural language attack on text classification
  and entailment.
\newblock In \emph{Proceedings of the AAAI Conference on Artificial
  Intelligence}.

\bibitem[{Kurita et~al.(2020)Kurita, Michel, and Neubig}]{kurita20acl}
Keita Kurita, Paul Michel, and Graham Neubig. 2020.
\newblock Weight poisoning attacks on pretrained models.
\newblock In \emph{ACL}.

\bibitem[{Li et~al.(2021{\natexlab{a}})Li, Zhang, Peng, Chen, Brockett, Sun,
  and Dolan}]{li2021CLARE}
Dianqi Li, Yizhe Zhang, Hao Peng, Liqun Chen, Chris Brockett, Ming-Ting Sun,
  and William~B Dolan. 2021{\natexlab{a}}.
\newblock Contextualized perturbation for textual adversarial attack.
\newblock In \emph{Proceedings of the 2021 Conference of the North American
  Chapter of the Association for Computational Linguistics}.

\bibitem[{Li et~al.(2023)Li, Shi, Liu, Kong, Wu, Zhang, Huang, and
  Lyu}]{li2023adversarial}
Guoyi Li, Bingkang Shi, Zongzhen Liu, Dehan Kong, Yulei Wu, Xiaodan Zhang,
  Longtao Huang, and Honglei Lyu. 2023.
\newblock \href {https://doi.org/10.18653/v1/2023.findings-emnlp.1053}
  {Adversarial text generation by search and learning}.
\newblock In \emph{Findings of the Association for Computational Linguistics:
  EMNLP 2023}, pages 15722--15738, Singapore. Association for Computational
  Linguistics.

\bibitem[{Li et~al.(2020)Li, Ma, Guo, Xue, and Qiu}]{li2020bertattack}
Linyang Li, Ruotian Ma, Qipeng Guo, Xiangyang Xue, and Xipeng Qiu. 2020.
\newblock Bert-attack: Adversarial attack against bert using bert.
\newblock In \emph{Proceedings of the Conference on Empirical Methods in
  Natural Language Processing and the International Joint Conference on Natural
  Language Processing}.

\bibitem[{Li et~al.(2021{\natexlab{b}})Li, Song, Li, Zeng, Ma, and
  Qiu}]{li2021backdoor}
Linyang Li, Demin Song, Xiaonan Li, Jiehang Zeng, Ruotian Ma, and Xipeng Qiu.
  2021{\natexlab{b}}.
\newblock \href {https://doi.org/10.18653/v1/2021.emnlp-main.241} {Backdoor
  attacks on pre-trained models by layerwise weight poisoning}.
\newblock In \emph{Proceedings of the 2021 Conference on Empirical Methods in
  Natural Language Processing}, pages 3023--3032, Online and Punta Cana,
  Dominican Republic. Association for Computational Linguistics.

\bibitem[{Liu et~al.(2019)Liu, Ott, Goyal, Du, Joshi, Chen, Levy, Lewis,
  Zettlemoyer, and Stoyanov}]{liu2019roberta}
Yinhan Liu, Myle Ott, Naman Goyal, Jingfei Du, Mandar Joshi, Danqi Chen, Omer
  Levy, Mike Lewis, Luke Zettlemoyer, and Veselin Stoyanov. 2019.
\newblock Roberta: A robustly optimized bert pretraining approach.
\newblock \emph{arXiv preprint arXiv:1907.11692}.

\bibitem[{Loshchilov and Hutter(2019)}]{loshchilov2017adamw}
Ilya Loshchilov and Frank Hutter. 2019.
\newblock Decoupled weight decay regularization.
\newblock In \emph{Proceedings of the International Conference on Learning
  Representations}.

\bibitem[{Pang and Lee(2005)}]{pang2005seeing}
Bo~Pang and Lillian Lee. 2005.
\newblock Seeing stars: Exploiting class relationships for sentiment
  categorization with respect to rating scales.
\newblock In \emph{Proceedings of the Annual Meeting of the Association for
  Computational Linguistics}.

\bibitem[{Qi et~al.(2021)Qi, Chen, Li, Yao, Liu, and Sun}]{qi2020onion}
Fanchao Qi, Yangyi Chen, Mukai Li, Yuan Yao, Zhiyuan Liu, and Maosong Sun.
  2021.
\newblock Onion: A simple and effective defense against textual backdoor
  attacks.
\newblock In \emph{EMNLP}.

\bibitem[{Sanh et~al.(2019)Sanh, Debut, Chaumond, and
  Wolf}]{sanh2019distilbert}
Victor Sanh, Lysandre Debut, Julien Chaumond, and Thomas Wolf. 2019.
\newblock Distilbert, a distilled version of bert: smaller, faster, cheaper and
  lighter.
\newblock \emph{5th NeurIPS Workshop on Energy Efficient Machine Learning and
  Cognitive Computing (2019), {ArXiv} preprint arXiv:1910.01108}.

\bibitem[{Torabi~Asr and Taboada(2019)}]{torabi2019big}
Fatemeh Torabi~Asr and Maite Taboada. 2019.
\newblock Big data and quality data for fake news and misinformation detection.
\newblock \emph{Big Data \& Society}, 6(1):2053951719843310.

\bibitem[{Wallace et~al.(2019)Wallace, Feng, Kandpal, Gardner, and
  Singh}]{wallace2019universal}
Eric Wallace, Shi Feng, Nikhil Kandpal, Matt Gardner, and Sameer Singh. 2019.
\newblock \href {https://doi.org/10.18653/v1/D19-1221} {Universal adversarial
  triggers for attacking and analyzing {NLP}}.
\newblock In \emph{Proceedings of the 2019 Conference on Empirical Methods in
  Natural Language Processing and the 9th International Joint Conference on
  Natural Language Processing (EMNLP-IJCNLP)}, pages 2153--2162, Hong Kong,
  China. Association for Computational Linguistics.

\bibitem[{Wang et~al.(2019)Wang, Singh, Michael, Hill, Levy, and
  Bowman}]{wang2018glue}
Alex Wang, Amanpreet Singh, Julian Michael, Felix Hill, Omer Levy, and Samuel~R
  Bowman. 2019.
\newblock Glue: A multi-task benchmark and analysis platform for natural
  language understanding.
\newblock In \emph{Proceedings of the International Conference on Learning
  Representations}.

\bibitem[{Wang et~al.(2021)Wang, Jin, Yang, and He}]{wang2021synonymencoding}
Xiaosen Wang, Hao Jin, Yichen Yang, and Kun He. 2021.
\newblock Natural language adversarial defense through synonym encoding.
\newblock In \emph{The Conference on Uncertainty in Artificial Intelligence}.

\bibitem[{Wu et~al.(2019)Wu, Morstatter, Carley, and
  Liu}]{wu2019misinformation}
Liang Wu, Fred Morstatter, Kathleen~M Carley, and Huan Liu. 2019.
\newblock Misinformation in social media: definition, manipulation, and
  detection.
\newblock \emph{ACM SIGKDD Explorations Newsletter}.

\bibitem[{Yan et~al.(2023)Yan, Gupta, and Ren}]{yan2023bite}
Jun Yan, Vansh Gupta, and Xiang Ren. 2023.
\newblock \href {https://doi.org/10.18653/v1/2023.acl-long.725} {{BITE}:
  Textual backdoor attacks with iterative trigger injection}.
\newblock In \emph{Proceedings of the 61st Annual Meeting of the Association
  for Computational Linguistics (Volume 1: Long Papers)}, pages 12951--12968,
  Toronto, Canada. Association for Computational Linguistics.

\bibitem[{Ye et~al.(2022)Ye, Miao, Wang, and Ma}]{ye2022texthoaxer}
Muchao Ye, Chenglin Miao, Ting Wang, and Fenglong Ma. 2022.
\newblock \href {https://doi.org/10.1609/aaai.v36i4.20303} {Texthoaxer:
  Budgeted hard-label adversarial attacks on text}.
\newblock \emph{Proceedings of the AAAI Conference on Artificial Intelligence},
  36(4):3877--3884.

\bibitem[{Yoo and Qi(2021)}]{yoo2021a2t}
Jin~Yong Yoo and Yanjun Qi. 2021.
\newblock Towards improving adversarial training of nlp models.
\newblock In \emph{Findings of EMNLP}.

\bibitem[{You et~al.(2023)You, Hammoudeh, and Lowd}]{you2023betteradversaries}
Wencong You, Zayd Hammoudeh, and Daniel Lowd. 2023.
\newblock \href {https://doi.org/10.18653/v1/2023.findings-emnlp.833} {Large
  language models are better adversaries: Exploring generative clean-label
  backdoor attacks against text classifiers}.
\newblock In \emph{Findings of the Association for Computational Linguistics:
  EMNLP 2023}, pages 12499--12527, Singapore. Association for Computational
  Linguistics.

\bibitem[{Zhan et~al.(2023)Zhan, Yang, Wang, Zheng, Huang, and
  Wang}]{zhan2023similarizing}
Pengwei Zhan, Jing Yang, He~Wang, Chao Zheng, Xiao Huang, and Liming Wang.
  2023.
\newblock \href {https://doi.org/10.18653/v1/2023.findings-acl.500}
  {Similarizing the influence of words with contrastive learning to defend
  word-level adversarial text attack}.
\newblock In \emph{Findings of the Association for Computational Linguistics:
  ACL 2023}, pages 7891--7906, Toronto, Canada. Association for Computational
  Linguistics.

\bibitem[{Zhan et~al.(2024)Zhan, Yang, Wang, Zheng, and
  Wang}]{zhan2024rethinking}
Pengwei Zhan, Jing Yang, He~Wang, Chao Zheng, and Liming Wang. 2024.
\newblock \href {https://aclanthology.org/2024.lrec-main.1223} {Rethinking
  word-level adversarial attack: The trade-off between efficiency,
  effectiveness, and imperceptibility}.
\newblock In \emph{Proceedings of the 2024 Joint International Conference on
  Computational Linguistics, Language Resources and Evaluation (LREC-COLING
  2024)}, pages 14037--14052, Torino, Italia. ELRA and ICCL.

\bibitem[{Zhang et~al.(2023{\natexlab{a}})Zhang, Huang, Wu, and
  Lyu}]{zhang2023semantics}
Jianping Zhang, Yung-Chieh Huang, Weibin Wu, and Michael~R. Lyu.
  2023{\natexlab{a}}.
\newblock \href {https://doi.org/10.24963/ijcai.2023/60} {Towards semantics-
  and domain-aware adversarial attacks}.
\newblock In \emph{Proceedings of the Thirty-Second International Joint
  Conference on Artificial Intelligence, {IJCAI-23}}, pages 536--544.
  International Joint Conferences on Artificial Intelligence Organization.
\newblock Main Track.

\bibitem[{Zhang et~al.(2023{\natexlab{b}})Zhang, Xiao, Li, Lv, Qi, Liu, Wang,
  Jiang, and Sun}]{zhang2021red}
Zhengyan Zhang, Guangxuan Xiao, Yongwei Li, Tian Lv, Fanchao Qi, Zhiyuan Liu,
  Yasheng Wang, Xin Jiang, and Maosong Sun. 2023{\natexlab{b}}.
\newblock \href {https://doi.org/10.1007/s11633-022-1377-5} {Red alarm for
  pre-trained models: Universal vulnerability to neuron-level backdoor
  attacks}.
\newblock \emph{Machine Intelligence Research}, 20(2):180--193.

\bibitem[{Zhou et~al.(2019)Zhou, Guan, Bhat, and Hsu}]{zhou2019fake}
Zhixuan Zhou, Huankang Guan, Meghana~Moorthy Bhat, and Justin Hsu. 2019.
\newblock \href {https://doi.org/10.5220/0007566307940800} {Fake news detection
  via {NLP} is vulnerable to adversarial attacks}.
\newblock In \emph{Proc. of the 11th Int. Conf. on Agents and Artificial
  Intelligence, {ICAART'19}}, pages 794--800.

\end{thebibliography}


\end{document}